\documentclass{bmvc2k}

\title{Two-Level Temporal Relation Model for Online Video Instance Segmentation}

\addauthor{Cagan Selim Coban}{ccoban20@ku.edu.tr}{1}
\addauthor{Oguzhan Keskin}{oguzhan.keskin@boun.edu.tr}{2}
\addauthor{Jordi Pont-Tuset}{jponttuset@google.com}{3}
\addauthor{Fatma Guney}{fguney@ku.edu.tr}{1}

\addinstitution{
 KUIS AI\\
 Koc University\\
 Istanbul, Turkey
}
\addinstitution{
 Bogazici University \\
 Istanbul, Turkey
}

\addinstitution{
    Google Research \\
    Zurich, Switzerland
}

\runninghead{COBAN et al.}{Two-Level Temporal Relation Model for online VIS}

\def\eg{\emph{e.g}\bmvaOneDot}

\newcommand{\bp}{\mathbf{p}}

\newcommand{\bz}{\mathbf{z}}

\newcommand{\bo}{\mathbf{o}}

\newcommand{\be}{\mathbf{e}}

\newcommand{\figref}[1]{\Fig~\ref{#1}}
\newcommand{\secref}[1]{Section~\ref{#1}}

\newcommand{\tabref}[1]{Table~\ref{#1}}

\makeatletter
\DeclareRobustCommand\onedot{\futurelet\@let@token\@onedot}
\def\@onedot{\ifx\@let@token.\else.\null\fi\xspace}
\def\eg{e.g\onedot} 
\def\ie{i.e\onedot} 
 
 \def\vs{vs\onedot}

\def\Fig{Fig\onedot}   
\makeatother

\newcommand{\xdownarrow}[1]{%
  {\left\downarrow\vbox to #1{}\right.\kern-\nulldelimiterspace}
}

\newcommand{\xuparrow}[1]{%
  {\left\uparrow\vbox to #1{}\right.\kern-\nulldelimiterspace}
}

\usepackage{tikz}
\def\checkmark{\tikz\fill[scale=0.4](0,.35) -- (.25,0) -- (1,.7) -- (.25,.15) -- cycle;}

\usepackage{colortbl}
\definecolor{mygray}{gray}{.92}
\definecolor{myred}{rgb}{1,0.5,0.5}
\definecolor{myred2}{rgb}{0.75,0,0}

\usepackage{multirow}
\usepackage{xcolor}

\begin{document}

\maketitle
\begin{abstract}
In Video Instance Segmentation (VIS), current approaches either focus on the quality of the results, by taking the whole video as input and processing it offline; or on speed, by handling it frame by frame at the cost of competitive performance. In this work, we propose an online method that is on par with the performance of the offline counterparts. We introduce a message-passing graph neural network that encodes objects and relates them through time. We additionally propose a novel module to fuse features from the feature pyramid network with residual connections. Our model, trained end-to-end, achieves state-of-the-art performance on the YouTube-VIS dataset within the online methods. Further experiments on DAVIS demonstrate the generalization capability of our model to the video object segmentation task. Code is available at: \url{https://github.com/caganselim/TLTM}
\end{abstract}

\section{Introduction}
\label{sec:intro}
Video Instance Segmentation (VIS) is the task of concurrently detecting, segmenting, and tracking object instances in videos. The recent progress in VIS is mainly driven by large datasets~\cite{yang2019video, xu18eccv, ponttuset17arxiv, caelles19arxiv, everingham2010pascal} that allow solving these tasks together. The existing methods can be categorized as \emph{offline}, when they take the whole video clip as input, or \emph{online}, when they process each frame or pair of frames sequentially.
Online methods typically follow the tracking-by-segmentation paradigm by first performing instance segmentation and then merging instances through time via an association algorithm~\cite{Yang_2021_ICCV,QueryInst,li2021spatial,liu2021sg,wu2021track,wang2021end2}. While frame-by-frame processing is fast, it lacks temporal context which results in a large number of ID switches due to e.g. occlusions. In contrast, offline methods are able to better leverage the spatio-temporal information from all the frames in the video~\cite{athar2020stem,bertasius2020classifying,lin2021video,vistr}. While temporal information leads to stronger performance, it hurts efficiency compared to the online counterparts, which might not be suitable for real-time applications.

In this paper, we propose an efficient online method that can reach the performance of offline methods. We achieve that by modeling the changes to the representation of the objects at two levels to allow the flow of temporal information between frames: a Graph Neural Network (GNN) at the object level and a spatio-temporal feature encoder, coined \textit{ResFuser}, at the feature level.
We build on top of a one-stage instance segmentation network, CenterMask~\cite{CenterMask} and
at the feature level, we aggregate features with the proposed ResFuser module, which uses a skip connection to predict changes from one frame to the next. 
At the object level, we estimate the changes to the states of objects through time and model the interactions between them with a GNN~\cite{kipf2019contrastive}: Each object becomes a node on the graph represented by its encoded state. The states of objects are updated via message passing between all the objects in consecutive frames. In the end, we associate the objects with an online tracker across multiple frames~\cite{yang2019video}.
Despite the two-level temporal aggregation, our network can be trained end-to-end.

Our comprehensive experiments on Youtube-VIS show the importance of both the feature level and the object level aggregation for the competitive results. Further experiments on DAVIS show our method’s generalization capability to one more task: Video Object Segmentation (VOS). We summarize our main contributions as follows:
\begin{itemize}
    \item \textbf{ResFuser for feature sharing in the Feature Pyramid Network (FPN)~\cite{lin2017feature} with residual connections:} This module enables to predict the changes to the feature maps of consecutive frames at each level of the FPN to relate the two frames at the feature-level.
    \item \textbf{A message-passing GNN that encodes objects and relates them across time:} This module exploits the fact that an object is still the most similar to itself from one frame to the next, despite some changes which typically occur due to the objects interacting with each other and with the environment. These interactions lead to some changes in the appearance of the object which we learn to estimate with a GNN.
\end{itemize}

\section{Related Work}
\label{sec:rw}
\paragraph{Video Instance Segmentation~(VIS).} 
Methods in VIS can be categorized as online or offline, depending on whether they operate \textit{frame-by-frame} or take the whole video as input.

\textbf{Online} methods typically add a \textit{tracking head} to a segmentation network such as Mask R-CNN~\cite{he2017mask}. In Mask-Track R-CNN~\cite{yang2019video}, tracking is formulated as a multi-class classification problem into one of the already identified instances or a new unseen instance. This methodology, called tracking-by-segmentation, extends an instance segmentation method with a matching algorithm and a history queue. QueryInst~\cite{QueryInst} treats instances as learnable queries with a multi-stage end-to-end framework and adopts the tracking method proposed in~\cite{yang2019video}. SipMask~\cite{cao2020sipmask} follows the same association strategy by utilizing a single-stage segmentation network. Single-stage methods for VIS have progressed quite significantly in recent years, leveraging spatio-temporal feature fusion techniques. CrossVIS~\cite{Yang_2021_ICCV} uses features of instances in a frame to localize the same instances in another frame at the pixel level. STMask~\cite{li2021spatial} employs a frame-level feature calibration module between predicted and ground-truth boxes and a temporal fusion module between consecutive frames to improve the inference of instance masks. SG-Net~\cite{liu2021sg} proposes a one-stage framework based on FCOS~\cite{FCOS} to improve mask quality by dividing target instances into sub-regions and performing instance segmentation on each sub-region. 

We propose an online method by using the single-stage CenterMask~\cite{CenterMask} as the instance segmentation network.
We extend the temporal context with two modules, ResFuser and GNN, at the feature and object levels, respectively. VisSTG~\cite{wang2021end2} also uses a GNN to relate the frames, but at the pixel level, which is too costly due to message passing between many pixels. This, in particular, entails that larger backbones cannot be utilized due to memory constraints. In contrast, our method relates objects with a GNN but still benefits from correlations at the feature level with the ResFuser, which is significantly less costly.

\textbf{Offline} methods operate on the whole video or on multiple frames to make better use of temporal information but this typically hinders efficiency and prevents their use in real-time scenarios. STEm-Seg~\cite{athar2020stem}, for example, models a video clip as a spatio-temporal volume where instances are represented as clusters inside the volume.
Instead of costly 3D volume processing, we use a single-stage segmentation network in the detection phase.
Based on DETR~\cite{carion2020end}, VisTR~\cite{vistr} uses transformers, but is not fully end-to-end trainable and has slow convergence as explained in~\cite{wu2022efficient}.
The performance of EfficientVIS~\cite{wu2022efficient} heavily depends on the number of input frames.
As shown in our experiments, our method taking just 2 frames as input outperforms the results of EfficientVIS with 9 frames.

\paragraph{Video Object Segmentation (VOS).} VOS is the task of segmenting and tracking arbitrary, novel objects without considering their semantic categories.
Depending on the input at test time, VOS can be divided into semi-supervised (or one-shot), where an initial mask of the object of interest is given~\cite{Man+18b,perazzi17cvpr,voigtlaender2019feelvos,wang19cvpr,wang19iccv}, and unsupervised (or zero-shot), without any initial mask~\cite{Lu_2019_CVPR,Ventura_2019_CVPR,wang19cvprb,zhou20tip,yang19iccv}.
Our method falls into the zero-shot category.
While some benchmarks consider only the dominant object in the scene for segmenting and tracking~\cite{perazzi17cvpr}, in the more generic case the videos have multiple objects~\cite{ponttuset17arxiv,xu18eccv,caelles19arxiv}.
AGNN~\cite{wang2019zero} builds a fully connected graph where nodes correspond to frames and a mask on each frame is related through message passing between them via an attention mechanism. While this works well for the case where there is a single foreground object moving in the video, we focus on the more realistic setting with multiple objects moving and interacting.

\section{Two-Level Temporal Relation Model}
\label{sec:method}
\begin{figure*}[t]
	\centering
   \includegraphics[width=\linewidth]{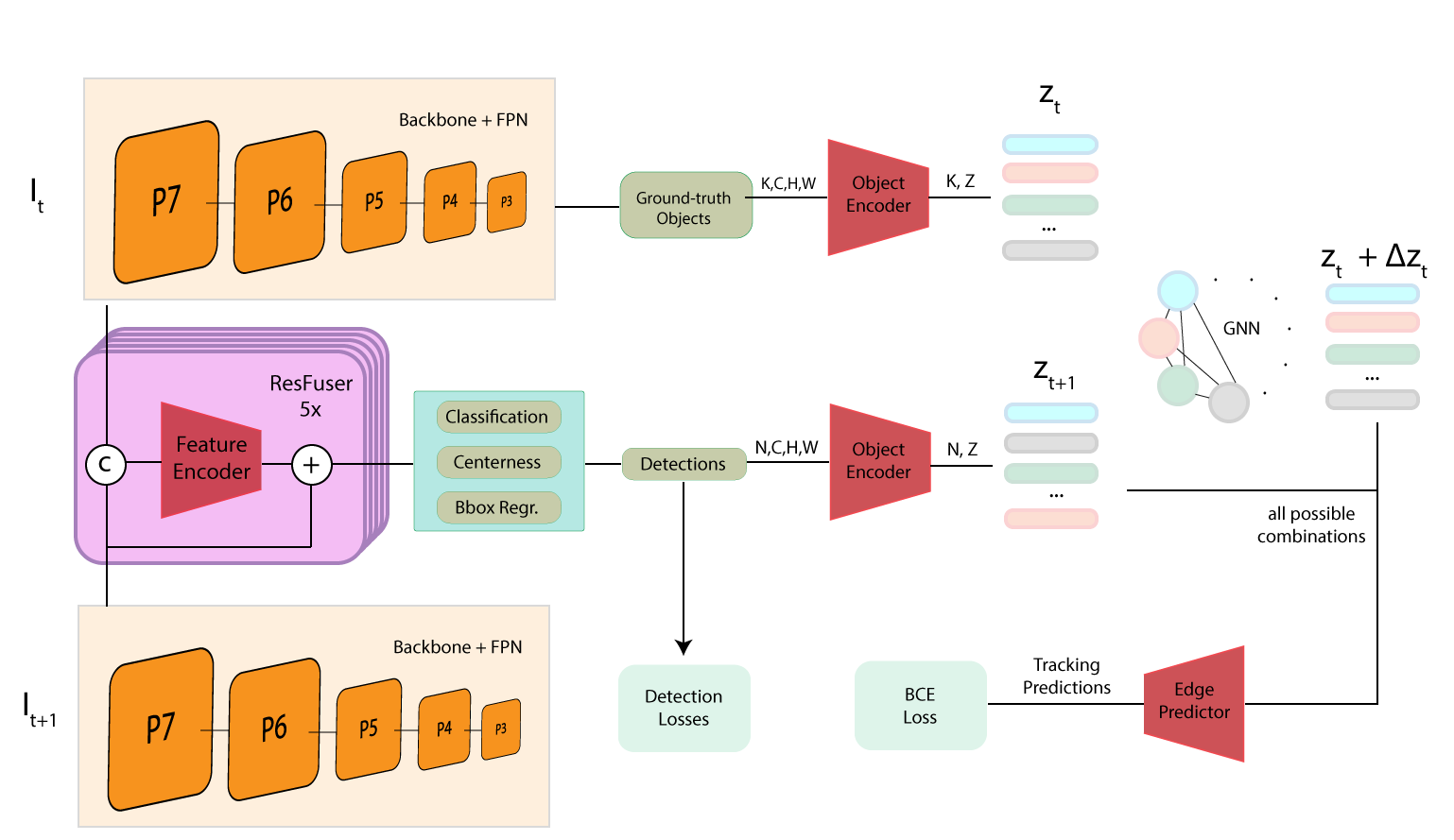}
	\caption{\textbf{Overall Framework.} Our method takes two consecutive frames as input and relates them temporally at two levels to better segment and track instances in videos: (i) feature level with ResFuser at each level of the feature pyramid and (ii) object level with a GNN after encoding objects with the object encoder.}
	\label{fig: para}
\end{figure*}
%
Our framework consists of two main parts: (i) we modify an instance segmentation model to aggregate spatio-temporal information between two consecutive frames and (ii) we use an object encoder to represent the objects and then relate them with a GNN-based transition model to learn changes to the representation of the objects. Below we explain the details of each part in detail.

\subsection{Spatio-Temporal Feature Aggregation (ResFuser)}
 
We start with an anchor-free image object segmentation method, CenterMask~\cite{CenterMask}, based on a single-stage object detector, FCOS~\cite{FCOS}. FCOS utilizes multi-level predictions with a Feature Pyramid Network~(FPN) to detect objects of various sizes.
 
Given an input frame at time $t$, FCOS extracts features $\bp_i^t$ at each level $i$. For spatio-temporal feature aggregation, we introduce the ResFuser module with the following functionality:
\begin{align}
    \bp_i^{(t+1)} = f_i(\bp_i^{(t)}, \bp_i^{(t+1)}) + \bp_i^{(t+1)}
\end{align}
where each $f_i$ is implemented as a two-layer CNN to learn residual changes at each level $i = \{3, \dots, 7\}$. The residual connections allow the fusing of missing information from the previous frame at each level.

A common approach in the design of segmentation networks is to add a separate head for segmentation in addition to the detection and the classification heads of an object detector, \eg as in Mask R-CNN~\cite{he2017mask}.
CenterMask adds a spatial attention-guided mask (SAG-Mask) branch to the anchor-free one-stage object detector FCOS~\cite{FCOS}. Given the bounding boxes from the detector, the SAG-Mask branch predicts a segmentation mask for each box with the spatial attention map. The goal of the attention map is to learn to focus on the pixels belonging to the object and learn to ignore the others around the object. The input to the SAG-Mask layer is the features inside the predicted region of interest (RoI) extracted by RoI Align~\cite{he2017mask}. 
Finally, the result of this operation is upsampled and used to predict class-specific masks. We use the same classification, regression, centerness, and mask losses as CenterMask.
We use the result of RoI Align also to represent objects as explained next.

\subsection{Encoding and Relating Objects}
\paragraph{Object Encoder.} Our object encoder takes the pooled object-centric features $\bo_t^k$ of CenterMask as input for each object $k$ at time $t$. These features are used to estimate the mask of the object but are high dimensional, \eg $256 \times 14 \times 14$ with a ResNet-50 (the number of channels times the size of the spatial grid).
We first encode these features into a low-dimensional latent representation $\bz^k_t$, for each object $k$ on both frames $t$ and $t+1$:
\begin{equation}
    \label{eq:enc_enc}
    \bz_t^k = \mathcal{E} (\bo_t^k)\qquad\bz_{t+1}^k = \mathcal{E} (\bo_{t+1}^k)
\end{equation}
with an object encoder $\mathcal{E}$ which is implemented as a simple two-layer CNN followed by an MLP. We relate the objects to each other in this latent representation as explained next.

\paragraph{Relating Objects.} Inspired by C-SWM~\cite{kipf2019contrastive}, we relate the objects using a Graph Neural Network (GNN). GNN models the state transitions of objects in the latent space by considering pairwise interactions between them. While we cannot explicitly model the action on the object which causes the transition as in C-SWM, we take advantage of an object being the most similar to itself from one frame to the next.

We construct a fully-connected graph withe the objects as nodes.
Our goal is to learn the state transitions, \ie the change in the state representation of each object on the graph from one frame to the next,
and we do so by passing messages between them.
The key observation is that the state transition of a node in the GNN should match the residual in between the object representation in two consecutive frames.
We use the same object encoder in the following frame and enforce the difference between the state representations of the first frame and the next to be the state transition estimated by the GNN.

We use the encoded latent representations $\bz_t^k$ as node features for each object $k \in \{1, \dots, K\}$ where $K$ is the number of objects segmented on frame $t$.
Our goal is to learn the transition $\Delta \bz_t^k$ to model the change from frame $t$ to $t+1$:
\begin{equation}
    \bz_{t+1}^k = \bz_t^k + \Delta \bz_t^k
\end{equation}
at each node/object $k$. The output of the GNN is the set of transitions for all $K$ objects at frame $t$, $\Delta \bz_t = \{\Delta \bz_t^1, \dots, \Delta \bz_t^K\}$.

While learning the transitions, GNN relates objects to each other by passing messages between them. The message passing operation is performed by iteratively updating the node representations:
\begin{align}
    \be_t^{i,j} = f_(\bz_t^i,~\bz_{t+1}^j) 
    && \hfill
    \Delta \bz_t^j = f_n(\bz_t^j,~\sum_{i \ne j} \be_t^{i, j})
\end{align}
where $f_n$ and $f_e$ are the node and the edge update functions, respectively; and $\be_t^{i,j}$ is the edge representation for the edge between the node $i$ and $j$ at time $t$. After one step of message passing, nodes become more aware of the other objects in the scene as well as their transitions. 

We train the network a binary-cross-entropy loss for associating instances in a video. We define a positive edge between the ground-truth instance node of an object in the first frame and a proposal node that belongs to the same object in the next frame. During inference, we employ the same tracking scheme as MaskTrack R-CNN~\cite{yang2019video} with the following modification on the score matrix: We use the predicted edge scores instead of the inner-product of object vectors as done in \cite{yang2019video}.

\section{Experiments}
\label{sec:exp}
\subsection{Training}
\label{sec:training}
\paragraph{Pre-training.} We first train CenterMask~\cite{CenterMask} on a combination of COCO~\cite{COCO}, YouTube-VIS~\cite{yang2019video} and Open Images~\cite{kuznetsova2020open} for the 40 classes of YouTube-VIS 2019, similarly to~\cite{luiten2020unovost}.
We used 19 overlapping classes of COCO and we map 60 similar classes of Open Images to the 40 classes of YouTube-VIS (see the Supplementary for the details of the mapping), \ie similar classes are represented with samples from both datasets. 

\paragraph{Training.} For data augmentation, we apply random affine transformations and motion blur as in~\cite{athar2020stem} to simulate video movement from static images.
We use the same losses as in CenterMask~\cite{CenterMask} except MaskIoU~\cite{huang2019mask}, which we have observed to add little to the performance (see the Supplementary for details).
We train our method using Stochastic Gradient Descent (SGD) for 180K iterations with a batch size of 16 and an initial learning rate of $5e^{-3}$.  We decrease the learning rate by a factor of 10 after 100K and 150K iterations. We will share the code and the pre-trained models upon publication. 

\begin{table*}[t!]
\begin{center}	
    \small
    \begin{tabular}{l|c|cc | cccc}
        \rowcolor{mygray}
        \hline
        \textbf{Method (ResNet-50)} & \textbf{Aug.} & \textbf{FPS} & \textbf{AP} & \textbf{AP50} & \textbf{AP75} & \textbf{AR1} & \textbf{AR10} \\  
        \hline
        MaskTrack R-CNN~\cite{yang2019video} & -  & 33 & 30.3 & 51.1 & 32.6 & 31.0 & 35.5 \\
        STEm-Seg~\cite{athar2020stem}& -  & 7 & 30.6 & 50.7 & 33.5 & 31.6 & 37.1 \\
        SipMask~\cite{cao2020sipmask} & -  & 34 & 32.5 & 53.0 & 33.3 & 33.5 & 38.9 \\
        CompFeat~\cite{fu2020compfeat} & - & $<$33 & 35.3 & 56.0 & 38.6 & 33.1 & 40.3  \\
        TraDeS~\cite{wu2021track} & -  & 26 & 32.6 & 52.6 & 32.8 & 29.1 & 36.6  \\
        CrossVIS~\cite{Yang_2021_ICCV} & - & 40 & 34.8 & 54.6 & 37.9 & 34.0 & 39.0 \\
        STMask~\cite{li2021spatial} &  DCN \cite{dai2017deformable} & 29 & 33.5 & 52.1 & 36.9 & 31.1 & 39.2\\
        SG-Net~\cite{liu2021sg}& MS & 23 & 34.8 & 56.1 & 36.8 & 35.8 & 40.8\\
        \textbf{Ours} & MS & 15 & 35.8 & 57.1 & 38.0 & 34.7 & 41.2 \\
        QueryInst~\cite{QueryInst} & MS & 32 & 36.2 & 56.7 & 39.7 & \bf 36.1 & \bf 42.9 \\
        CrossVIS~\cite{Yang_2021_ICCV}& MS  & 40 & 36.3 & 56.8 & 38.9 & 35.6 & 40.7\\
        VisSTG~\cite{wang2021end2} & MS & 22 & \bf 36.5 & \bf 58.6 & \bf 39.0 &  35.5 & 40.8\\
        \hline
    \end{tabular}\\[2mm]
    \begin{tabular}{l|c|cc | cccc}
        \hline
        \rowcolor{mygray}
        \textbf{Method (ResNet-101)} & \textbf{Aug.} & \textbf{FPS} & \textbf{AP} & \textbf{AP50} & \textbf{AP75} & \textbf{AR1} & \textbf{AR10} \\ 
        \hline
        MaskTrack R-CNN~\cite{yang2019video} &  -  & 29 & 31.9 & 53.7 & 32.3 & 32.5 & 37.7\\
        SRNet~\cite{ying2021srnet} & - & 35 & 32.3 & 50.2 & 34.8 & 32.3 & 40.1\\
        STEm-Seg~\cite{athar2020stem} & -  & 7 & 34.6 & 55.8 & 37.9 & 34.4 & 41.6 \\
        CrossVIS~\cite{Yang_2021_ICCV} & -  & 36 & 36.6 & 57.3 & 39.7 & 36.0 & 42.0\\			
        SipMask~\cite{cao2020sipmask}& MS & 24 & 35.8 & 56.0 & 39.0 & 35.4 & 42.4\\
        STMask~\cite{li2021spatial}& DCN \cite{dai2017deformable} & 23 & 36.8 & 56.8 & 38.0 & 34.8 & 41.8 \\
        SG-Net~\cite{liu2021sg}& MS & 20 & 36.3 & 57.1 & 39.6 & 35.9 & 43.0 \\
        \textbf{Ours}& MS & 14 & \bf 38.1 & \bf 61.9 & \bf 40.6 & \bf 37.0 & \bf 44.4\\
        \hline
    \end{tabular}
\end{center}
\caption{\textbf{Results on YouTube-VIS 2019.} We compare our method to other online state-of-the-art methods on YouTube-VIS 2019 \texttt{val} set using ResNet-50 (top) and ResNet-101 (bottom) as backbones. ``MS'' denotes multi-scale training.} 
\label{tab:ytvis19}
\end{table*}

\subsection{Benchmarks}
\paragraph{YouTube-VIS 2019~\cite{yang2019video}.} The YouTube Video Instance Segmentation 2019 dataset is the first large benchmark introduced for the VIS task.
It consists of 2883 high-quality YouTube videos containing 4883 unique objects annotated with approximately 131k object masks corresponding to 40 predetermined object categories. The task is to segment and classify each object instance while consistently tracking them across frames. The evaluation metrics are Average Precision (AP) and Average Recall (AR). These metrics are evaluated over 10 IoU thresholds from 50\% to 95\% at a step of 5\%, and calculated firstly by category and finally averaged over the category set.

\paragraph{DAVIS 2019~\cite{caelles19arxiv}.} Densely Annotated Video Segmentation 2019 is a high-quality dataset that is popular in the VOS task. In its 2019 version, there are a total of 90 sequences, 60 for training and 30 for validation. We evaluate our framework on the Unsupervised Video Object Segmentation (UVOS) task, \ie to segment and track foreground objects without classifying them, unlike VIS.
The main evaluation metric is the $J\&F$ score, which is the mean of $J$-score and $F$-score. The former is based on the IoU between the predicted and the ground masks whereas $F$-score measures the accuracy of the predicted mask boundaries.

\subsection{Quantitative Results}
\subsubsection{Video Instance Segmentation}
\tabref{tab:ytvis19} compares our method to online and real-time state-of-the-art VIS methods on YouTube-VIS 2019. Following the literature, we perform a separate comparison by using a small~(ResNet-50 \cite{he2016deep}) and a large~(ResNet-101 \cite{he2016deep}) backbone. The results show that our model outperforms the other models by 1.8 points with the ResNet-101 backbone while preserving a competitive performance with ResNet-50.
.
Our model extends the capability of online models to benefit more from the temporal context with the two proposed modules, GNN and ResFuser. Furthermore, our model can better utilize a large number of parameters as shown in the SoTA performance with the large backbone. Since these two modules have additional parameters, more data is required, both in terms of quantity and variety as shown in our ablations~(\secref{sec:ablation}).
The only other model that employs a GNN for feature aggregation is VisSTG~\cite{wang2021end2} but their results are not reported with the ResNet-101 backbone, probably due to overly increased computational and memory requirements.

Note that offline methods which make use of the full sequence are not included in \tabref{tab:ytvis19} for a fair comparison. For reference, VisTR~\cite{vistr}, a real-time and end-to-end trainable encoder-decoder transformer-based framework, reports a score of 35.6 mAP with ResNet-50 backbone and 38.6 mAP with ResNet-101 backbone with $T=36$ frames as input, which is the largest amount of annotated frames on YouTube-VIS-19 dataset. Similarly, the state-of-the-art EfficientVIS~\cite{wu2022efficient} achieves 37.9 mAP with ResNet-50 and 39.8 mAP with ResNet-101 backbone using $T=36$. These high scores can be attributed to rich information available on the whole video level as proven by a score of 35.3 mAP when reducing the temporal window to $T=9$ input frames~\cite{wu2022efficient}, while our method can obtain 35.8 mAP with just $T=2$ frames.
Although STEm-Seg~\cite{athar2020stem} is also a whole-video-level method, we still include it in our comparisons because, to the best of our knowledge, it is the first method that reports results on both VIS and VOS tasks like us. 

\subsubsection{Video Object Segmentation}
\begin{table*}[t!]
\resizebox{\textwidth}{!}{%
    \begin{tabular}{l|c|c|c|ccc|ccc}
        \rowcolor{mygray}
        \hline
        \textbf{Method} & \textbf{Online} & \textbf{E2E} & \textbf{J \& F} & \textbf{Mean} & \textbf{Recall} & \textbf{Decay} & \textbf{Mean} & \textbf{Recall} & \textbf{Decay}  \\ 
        \hline
        RVOS~\cite{Ventura_2019_CVPR} & \multirow{3}{*}{\checkmark} & \checkmark & 41.2  & 36.8 & 40.2 & 0.5 & 45.7 & 46.4 & \textbf{1.7} \\
        KIS~\cite{cho2019key} &  & -  &  59.9 & - & - & - & - & - & - \\
        \textbf{Ours} &  & \checkmark &  \textbf{61.9} & \textbf{60.7} & \textbf{70.3} & \textbf{-2.4} & \textbf{63.1} & \textbf{70.9} & 1.9 \\ \hline
        AGNN~\cite{wang2019zero} & \multirow{4}{*}{-} &\ -  & 61.1 & 58.9 & 65.7 & 11.7 & 63.2 & 67.1 & 14.3 \\
        STEm-Seg~\cite{athar2020stem} &  & \checkmark   & 64.7 & 61.5 & 70.4 & \textbf{-4} & 67.8 & 75.5 & 1.2 \\
        UnOVOST~\cite{luiten2020unovost} &  & - &  67.9 & \textbf{66.4} & \textbf{76.4} & -0.2 & 69.3 & \textbf{76.9} & \textbf{0.01} \\      Propose-Reduce~\cite{lin2021video} &  & \checkmark & \textbf{68.3} & 65.0 & - & - & \textbf{71.6} & - & - \\
        \hline
    \end{tabular}}\\
\caption{\textbf{Results on DAVIS-19.} We compare our method to state-of-the-art methods, online at the top and offline at the bottom, on DAVIS-19 Unsupervised \texttt{val} set. All results are reported with the ResNet-101 backbone.}
\label{tab:davis19}
\end{table*}
We evaluate our model's performance on the additional task of VOS in \tabref{tab:davis19} without training on it.
Our method considerably outperforms previous online methods RVOS~\cite{Ventura_2019_CVPR} and KIS~\cite{cho2019key}, and achieves a slightly better $J\&F$ score with a noticeably less decay than offline AGNN~\cite{wang2019zero}.
Our method is only 2.5 points below the offline STEm-Seg, despite the 14-frame video input used by their method. With $T=4$ input frames, STEm-Seg reports a $J\&F$ score of 62.2 which is similar to our method's performance (61.9) with only $T=2$ input frames.
Furthermore, STEm-Seg is trained on both image (COCO~\cite{COCO} and PASCAL~\cite{everingham2010pascal}) and video datasets (YouTube-VIS and DAVIS) whereas our method is not even trained on DAVIS, which shows a better generalization performance to the video object segmentation task.
Although the offline UnOVOST~\cite{luiten2020unovost} method achieves an impressive score of 67.9, it cannot be trained end-to-end due to its complex post-processing based on heuristics that are specific to this benchmark.
Also, it is significantly slower (1 \vs 14 FPS).
Propose-Reduce~\cite{lin2021video}, current state of the art on DAVIS-19, is another offline method based on performing segmentation on keyframes and then propagating it with a two-stage network. Both the performance and efficiency are highly dependent on the number of keyframes.

\subsection{Ablation Study}
\label{sec:ablation}
\begin{table}[t!]
\begin{center}	
    \small
    \begin{tabular}{l|c|cc}
        \rowcolor{mygray}
        \hline
          & \textbf{YTVIS-19 (mAP)} & \textbf{DAVIS-19 (J\&F)} \\
          \hline
        No Open Images & 28.0 & 54.7 \\
        Open Images & 34.3 & 59.1 \\
        Open Images + ResFuser & 34.9 & 59.4 \\
        Open Images + ResFuser + GNN & \textbf{35.8} & \textbf{60.5} \\
        \hline
    \end{tabular}
\end{center}
\caption{\textbf{Ablation Study.} We show the effect of pre-training on OpenImages~\cite{kuznetsova2020open}, and our two contributions, ResFuser and GNN on YoutubeVIS and DAVIS. All results are presented with the ResNet-50 backbone.}  
\label{tab:ablation}
\end{table}
\tabref{tab:ablation} presents our ablation experiments to isolate the effect of each of our contributions.
The first row is a baseline with CenterMask pre-trained only on YouTube-VIS-19~\cite{yang2019video} and COCO~\cite{COCO}. Here, we use an IoU-based tracker by removing the Kalman Filter from the well-known SORT algorithm~\cite{bewley2016simple}. 
In the second row, we pre-train CenterMask~\cite{CenterMask} additionally on Open Images~\cite{kuznetsova2020open} to see the effect of a larger-scale image instance segmentation dataset on video segmentation problems. Although Open Images is not typically used for video segmentation problems, our framework greatly benefits from this additional data as it relies on instance detection and segmentation on the frame level.
We observe a significant boost of +6.3 mAP on YouTube-VIS-19 and +4.4 $J\&F$ on DAVIS-19.

We then add our feature-sharing module ResFuser which is a simple feature-sharing procedure between the consecutive frames to improve detection and segmentation by exploiting similar feature representations of corresponding instances.
Despite being conceptually very simple, ResFuser consistently improves the results on both datasets too (+0.6 mAP on YouTube-VIS-19 and +0.3 $J\&F$ on DAVIS-19).

Finally, the last row shows that linking objects across time with a GNN further improves the performance by a noticeable margin (+0.9 mAP in YouTube-VIS-19). The gain on the DAVIS-19 unsupervised validation set is even higher (+1.1 $J\&F$). This reinforces our claim that GNN enhances instance segmentation and tracking in multiple objects, which is the case for DAVIS-19 as it contains a large number of sequences with 2 or 3 objects (the average number of objects per sequence is 2.2 on the validation set~\cite{caelles19arxiv}). Our method of relating objects with a GNN shows promising improvements for both video segmentation tasks.

\subsection{Qualitative Analysis}
We show two qualitative examples in \figref{fig: visual} (see Supplementary for the videos) comparing the results with our GNN module versus those without.
In the case of VIS, our model with GNN can correctly segment and track the tennis racket and the two players despite frequent exiting and re-entering throughout the video. The model without GNN, on the other hand, assigns a different ID to each object when they re-enter the view, showing the benefits of our object-level reasoning in the GNN.

In the VOS video, we demonstrate our model's capability to segment and track object instances with similar appearances and frequent occlusions. The two fish occluding each other on the first frame (1 and 5) can be separated correctly with the help of GNN while they are treated as a single object without GNN. In the next frame, a fish makes a fast turn and in the third frame, another fish swims at the bottom edge of the frame. These situations significantly affect object appearances, and GNN continues to detect and track these fish as opposed to the model without GNN. Finally, in the last frame, our model with GNN can segment and identify the fish entering the frame (6) as a new object despite being heavily camouflaged. 

\begin{figure*}[t]
	\centering
   \includegraphics[width=\linewidth]{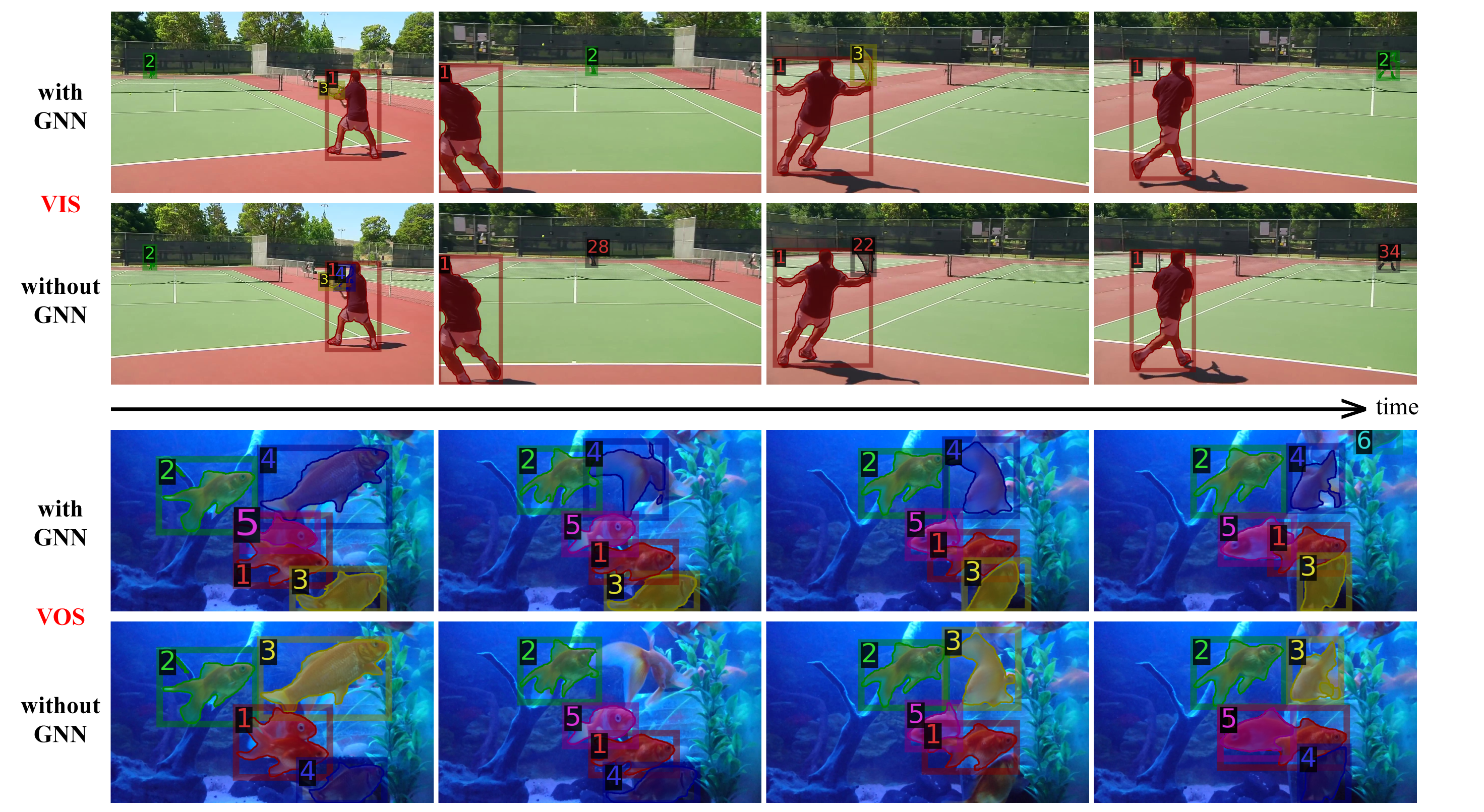}
	\caption{\textbf{Qualitative Results.}}
	\label{fig: visual}
\end{figure*}
\section{Conclusion}
\label{sec:conc}
We proposed a new end-to-end trainable framework for simultaneously detecting, segmenting, classifying, and tracking objects in videos in an online fashion. We introduced two key novelties: an effective skip-connection module (ResFuser) that allows feature sharing between consecutive frames and an object-relating GNN. The two together reach state-of-the-art results on the challenging YouTube-VIS dataset, with an AP of 38.1 at 14 frames per second. We showed that our framework generalizes well to the VOS task by achieving competitive results without ever seeing the corresponding training data.

\newpage
\bibliography{bibliography_long, egbib}

\end{document}